\documentclass[11pt]{article}

\usepackage[final]{acl}

\usepackage{times}
\usepackage{latexsym}

\usepackage[T1]{fontenc}

\usepackage[utf8]{inputenc}

\usepackage{microtype}

\usepackage{inconsolata}

\usepackage{graphicx}

\usepackage{supertabular}
\usepackage{booktabs}
\usepackage{xspace}
\usepackage[svgnames]{xcolor}

\title{Artwork Interpretation with Vision Language Models: \\
  A Case Study on Emotions and Emotion Symbols}

\author{Sebastian Padó$^*$ \and Kerstin Thomas$^{\dagger}$ \\
  $^*$ Institute for Natural Language Processing (IMS) \\
  $^{\dagger}$ Institute for Art History (IKG) \\
  University of Stuttgart, Germany \\
  \texttt{\{sebastian.pado@ims,kerstin.thomas@ikg\}.uni-stuttgart.de}}

\begin{document}
\maketitle
\begin{abstract}
  Emotions are a fundamental aspect of artistic expression. Due to
  their abstract nature, there is a broad spectrum of emotion
  realization in artworks. These are subject to historical change and
  their analysis requires expertise in art history.
  In this article, we investigate which aspects of emotional
  expression can be detected by current (2025) vision language models
  (VLMs). We present a case study of three VLMs (Llava-Llama and two
  Qwen models) in which we ask these models four sets of questions of
  increasing complexity about artworks (general content, emotional
  content, expression of emotions, and emotion symbols). We find that
  the VLMs recognize the content of the images surprisingly well and
  often also which emotions they depict and how they are expressed.
  The models perform best for concrete images but fail for highly
  abstract or highly symbolic images. Reliable recognition of symbols
  remains fundamentally difficult. Furthermore, the models continue to
  exhibit the well-known LLM weakness of providing inconsistent
  answers to related questions.
\end{abstract}

\section{Introduction}

Emotions are a privileged aspect of artistic expression in the visual
arts \cite{Tan2000EmotionArt}. Arguably, many artworks actually
address the emotions of the viewer far more directly than the
intellect, in order to gain privileged access to viewer's morals,
beliefs, values, and worldview. In fact, for a long time, the
recommendations for the art of persuasive speech (rhetoric) and those
for the visual arts have been directly related
\cite{Barthes1977-BARIMT-3}.

In recent years, digital access to artwork has scaled up considerably
and has become an important branch of (digital) cultural heritage
curation \cite{Naeslund01102020}. For example, Prometheus
\cite{Dieckmann2010Prometheus} is a distributed archive which as of
now (August 2025) provides a unified interface to access almost 4
million images from art, culture, and history. Such archives offer the
opportunity for 'scalable reading' \cite{Weitin2017ScalableReading} --
or, in this case, more accurately 'scalable viewing' -- studies in
which automated analysis on large data sets is combined with a focused
manual analysis on smaller samples and which have become widely used
in other areas of Digital Humanities, notably literary studies
\cite{saPereira2019mixed}. This approach would be particularly
attractive for research questions that inherently involve large number
of images, such as: Which elements in images have constant
significance for emotional expression, and which exhibit variance? How
have such elements changed historically? What cultural differences are
there in emotional expression?

In the textual modality -- also prominent in cultural heritage -- the
analysis of emotions has taken major steps in the last ten
years. Neural approaches can detect emotions significantly more
accurately than previous approaches \cite{nandwani21}. They no longer
rely on dictionaries of keywords but can recognize emotional contexts,
including stylistic devices such as irony, while also taking global
document meaning into account. Such models have also been used
successfully for the analysis of emotional content in literary texts
\cite{Kim2019SurveyLit}. Recently, vision language models (VLMs)
generalize the success of text-based methods by tightly integrating
language and image information, enabling cross-modal information
transfer: They enable the generation of images from textual
descriptions, conversely the generation of textual descriptions for
images, and the answering of textual questions on images
\cite{Antol2015VQA}.

This development has the potential to carry out emotion analysis on
artwork with VLMs. However, the success of this approach is all but
guaranteed.  Due to their abstract nature, emotions are realized
visually in a wide variety of ways. Artists can draw on the study of
natural emotional expression, on facial expressions and
gestures. Alternatively, an emotional expressions can result from
various abstract elements, such as colors, shapes, compositional
schemes, or symbols: signs with a secondary conventionalized level of
meaning \cite{cassirer}. Since their meaning can only be analyzed in
context and is subject to historical change, the analysis of emotional
content of artworks is generally assumed to require art historical
expertise. Additionally, on the technical level, it is well known that
VLMs, despite a surprisingly good understanding of individual aspects,
struggle to develop a globally coherent understanding and can suffer
from hallucinations \cite{Huang2025HallucinationSurvey}. Substantive
studies on (properties of) artworks however have to assume that the
automatic analysis is largely accurate, or that remaining errors are
at least distributed as randomly as possible.

To our knowledge, there are no studies that gauge the concrete quality
of emotion-related analyses that fall out of current VLMs. Our study
aims to fill this gap. We select 38 images of artworks (mostly of
paintings, but also including sculptures and photographs), present
these images to three current vision-language models and ask them a
catalog of eight questions, ranging from a pure description of the
content to the interpretation of the emotions and any symbols used. We
qualitatively evaluate the models' outputs by hand to understand how
differentiated the models generally perceive the different aspects of
our images.

We obtain mixed results: VLMs are capable of correctly recognizing
image content, often including emotions and the artistic means used to
express them. However, recognition generally seems to be based on
conventionalized patterns and fails with novel combinations. Complex
and symbolic images also pose difficulties. Our conclusion is that
current VLMs are already sufficiently accurate for some scalable
reading research questions and for some types of images, but not yet
for others.

\section{Background and Related Work}

\subsection{Emotions in Art History}

From an art historical perspective, emotions are central to the
artistic impact. A work of art is meant to persuade, move, and
stimulate thought. However, how works of art achieve this effect is
often far from obvious.

One possibility is for them to use means of expression from natural
life that are based on anthropological patterns: an open mouth with
raised corners indicates laughter, while narrowed eyes and drooping
corners indicate sadness. The same applies to physical means of
expression such as posture and gestures. The expressive values are
modular: one expressive value is reinforced by another. They are also
quite constant over time within a cultural area. For this reason, they
are referred to in psychology as 'basic emotions'
\cite{Ekman1999BasicEmotions}.
 
As an alternative to means of expression based on gestures and facial
expressions, artists employ less clearly legible means of expression
in their works, such as certain colors, shapes, compositional devices,
or symbols -- e.g., heart for love. These act alongside the
anthropological means (such as smiles), reinforcing them and leading
to specific emotional expressions. The knowledge needed to understand
these means is culturally and historically specific, and reading them
correctly requires a historical understanding. For this reason, many
researchers call for a model of 'historical emotion research' in
addition to the more universal model of basic emotions
\cite{Stearns1985Emotionology,Rosenwein2010ProblemsEmotions,Frevert2011Gefuehlswissen,Matt2011EmotionHistory,Plamper2012GeschichteGefuehl}.

Art historical research rarely distinguishes these two
scenarios. Instead, emotions in works of art are attributed
predominantly on the basis of concrete, codified expression schemes
and symbols. The more abstract means of conveying emotions are rarely
analyzed in detail, but rather take a back seat in the shape of
general and diffuse descriptions of impressions. Our study takes steps
towards remedying this situation, using VLMs as a pre-theoretical
device that gives -- at least in theory -- equal importance to the
different elements in the image. In practice, of course, this depends
on the materials on which the models we use were pretrained.

\subsection{Language and Vision-Language Models}

Attempts to support the analysis of artworks with automatic methods
can at this point build on the technological progress in AI/NLP of the
last ten year, notably Language Models (LM) based on transformers
\cite{Vaswani2017Attention}. In contrast to earlier approaches,
transformer-based LMs can consider an extensive linguistic context and
can be (pre-)trained on large datasets since their training
parallelizes well. From a user perspective, an important development
in recent years is the emergence of instruction-tuned models
\cite{Brown2020GPT3} that are able to answer textual questions without
task-specific training, which makes them directly applicable for
interactive text-based querying by domain experts.

Vision-Language Models (VLMs) go beyond language and strive to develop
a shared understanding of information from multiple modalities --
typically text and visual data. Conceptually, this is achieved by
bringing together embeddings from both modalities. Most VLMs consist
of an encoder for the image, whose output is projected onto the
embedding space of a language model that represents text
meaning. Early models such as VilBERT \cite{Lu2019ViLBERT} still use
conventional neural networks such as CNNs for the visual encoders.
Current models achieve a new level of quality by using contrastively
trained image encoders such as CLIP \cite{Radford2021CLIP}, trained to
match very large sets of image-description pairs. The resulting models
capture semantic concepts across domains. VLMs inherit the ability to
accept textual instructions and questions and generate verbal output
from their component LMs, enabling them to perform 'visual question
answering' \cite{Antol2015VQA}.

However, current models are clearly not perfect. Like LMs, VLMs
exhibit a tendency toward hallucinations
\cite{Liu2024HallucinationVLM}, which can be described as a
cross-modal inconsistency between image and description. Judging from
experience with LMs, we might expect that VLMs will have more
difficulty the less experience they have with a type of image, a type
of question, or both.

\subsection{Artwork Analysis with Language-Vision Models}

To our knowledge, there are only two previous studies in NLP that
study the outputs of VLMs for
artworks. \citet{hayashi-etal-2024-towards} create a corpus pairing
artworks with their corresponding Wikipedia articles. They ask VLMs to
generate texts corresponding to article parts (sections, subsections,
etc.) and evaluate the output against the actual article parts with
natural language generation metrics, both in terms of textual overlap
(e.g., BLUE and ROUGE) and in terms of entity
match metrics. \citet{ozaki-etal-2025-towards} extend this paradigm to
multiple languages and in addition assess the capability of tuning to
improve the models' outputs. These studies found somewhat mixed
results: The VLMs generally used a certain amount of ground-truth
entities, and LoRA tuning improved results, but this was only true for
English, and performance for other languages was lower throughout.

\section{Experimental Setup}

\subsection{Motivation}

In our study, we use the same fundamental generation paradigm as
\citet{ozaki-etal-2025-towards} -- prompting VLMs with a pair of
picture and question -- but combine it with a different evaluation
paradigm.  We do not evaluate the generated texts against a ground
truth, for two reasons: (a) NLG metrics, such as entity coverage, are
difficult to translate into insights, and it is hard to understand
from the previous studies how well the VLMs really understand the
images; (b) the questions are determined by the Wikipedia article
structures and thus include questions that are not visual in nature
('What is the history of the Mona Lisa?') while they do not include
topic-specific questions -- such as those concerning our topic of
interest, emotions.

To address concern (a), to carry out a qualitative analysis,
presenting the VLM outputs to two experts (the authors) to annotate
for reasonableness.\footnote{The quality assessments largely
  correspond, so the results we report below represent a consensus
  between the authors.} We thus trade a fully evaluation procedure
against a (hopefully) more detailed understanding of the
capabilities and limits of the VLMs. 

To realize this potential, we address concern (b) by defining a set of
eight questions that we ask the VLMs for each image, listed in
Table~\ref{tab:questions}. We formulate the questions in English,
since current LLMs and VLMs are trained predominantly with English
data \cite{Zhang2024VLSurvey}. Therefore, model errors observed for
English prompts can be more reliably interpreted as deficits in
conceptual understanding than linguistic shortcomings
\cite{ozaki-etal-2025-towards,Qin2025MultilingualLLM}.

\begin{table}[tb!h]
  \centering
\begin{tabular}{lp{0.80\columnwidth}}
    \toprule
 Q~1 &	Is this a painting, a drawing, a sculpture, or something else? Be brief. \\
 Q~2 &  	What does this artwork show? Be brief. \\ \midrule
Q~3 &  	Does this artwork involve an emotion? Give a yes/no answer and a brief justification. \\
Q~4 &  	Is the emotion shown by this artwork a positive or a negative one? Give a one-word answer (positive/negative) and a brief justification. \\
Q~5 & 	What emotion is shown by this artwork? Be brief. \\ \midrule
Q~6 & 	How is the emotion shown by this artwork expressed artistically? Do not write more than a short paragraph. \\
Q~7 & 	Does this artwork use a symbol to visualize emotion, and if it does, how? Do not write more than a short paragraph. \\
Q~8 & 	How intense is the emotion depicted? Be brief. \\ \bottomrule
  \end{tabular}
  \caption{Questions posed to VLMS for image interpretation: Basic description (Q 1--2),
    Emotion recognition (Q 3--5), Emotion expression (Q 6--8)}
  \label{tab:questions}
\end{table}

The questions progress from basic descriptions of form and content
(Q~1 and 2) to the interpretation of the displayed emotions (Q~3 to 5)
and more nuanced characteristics of these emotions, such as the means
used for representation (Q~6), the use of symbols (Q~7), and their
intensity (Q~8). The recurring request for brevity was necessary to
neutralize the models' well-known tendency to give lengthy answers
\cite{Wang2024CoT}.

We also note that our case study is focused on understanding the
visual understanding of the models in their default as-published
state. For this reason, we do not provide the images' titles to the
VLMs \cite{hayashi-etal-2024-towards} -- so that all information has
to be inferred from visual information -- and we do not experiment
with fine-tuning the VLMs \cite{ozaki-etal-2025-towards}.

\subsection{Selection of VLMs}

The largest and best-performing LLMs and VLMs are all proprietary and
can only be used via company-controlled APIs, which is problematic
from the perspective of transparency and reproducibility
\cite{Liesenfeld2023OpeningChatGPT}. We therefore limit our study to
three comparatively small VLMs that have open weights, i.e., can be
downloaded in their entirety, and are small enough to be executed
locally on a single 48GB GPU core.

\newcommand{\llava}{\textsc{Llava-Llama-8b}\xspace}
\newcommand{\qwen}{\textsc{Qwen-7b}\xspace}
\newcommand{\qwenawq}{\textsc{Qwen-32b-awq}\xspace}

Our first model is \llava. It is a member of the Llava 'Large Language
and Vision Assistant' \cite{Liu2023VisualInstruction} model family. It
combines CLIP as an image encoder with Meta AI's large multilingual
Llama 8B LM. The two other models are variants of the Alibaba Qwen-VL
architecture \cite{Bai2025Qwen25VL} which combines a CLIP image
encoder with the Qwen 2.5 multilingual LMs. We use a version with 7B
parameters (\qwen) and one with 32B parameters quantized with AWQ
(\qwenawq, \citet{MLSYS2024_42a452cb}).

The three models are broadly similar in that they use similarly
structured image encoders as well as similar transformer-based
LMs. However, they behave significantly differently in practice, which
is due to the fact that they were trained by different developers on
different data. Unfortunately, we do not have sufficiently detailed
information about the training process of any of the models to predict
specific behavioral patterns.

\subsection{Selection of Images}

We compile a set of 38 images of artworks from the Prometheus image
archive \cite{Dieckmann2010Prometheus}, a huge data base of (among
other domains) artworks from over 100 different organizations. We
select artworks from a wide range of periods (including the Middle
Ages, Renaissance, Baroque, and the 18th to 20th centuries) that have
relatively clear emotional content. Our sample comprises mostly
paintings, supplemented by some drawings, photographs, and images of
sculptures. We always select artworks in minimal pairs of two similar
images to gauge the robustness of interpretation. The complete image
list can be found in Appendix A.

We present most image files directly as downloaded from
Prometheus. For two paintings by Arnold Böcklin, irrelevant parts of
the two images are masked out. In two other cases, recognizable
writing on the images is masked to focus the model on visual
information (see Appendix A for details).

The images are presented to the VLMs individually with each question.
We set maximum answer length to 150 tokens and otherwise use the
models' default parameters. This results in a total of 20 image pairs
x 2 images x 8 questions x 3 models = 960 answers.\footnote{The
  metadata of the image files, the responses of the models, and the
  scripts used for the evaluation are publicly available at
  \url{https://doi.org/10.5281/zenodo.17668625}}

\section{Findings}

\begin{figure*}
  \includegraphics[width=0.38\textwidth]{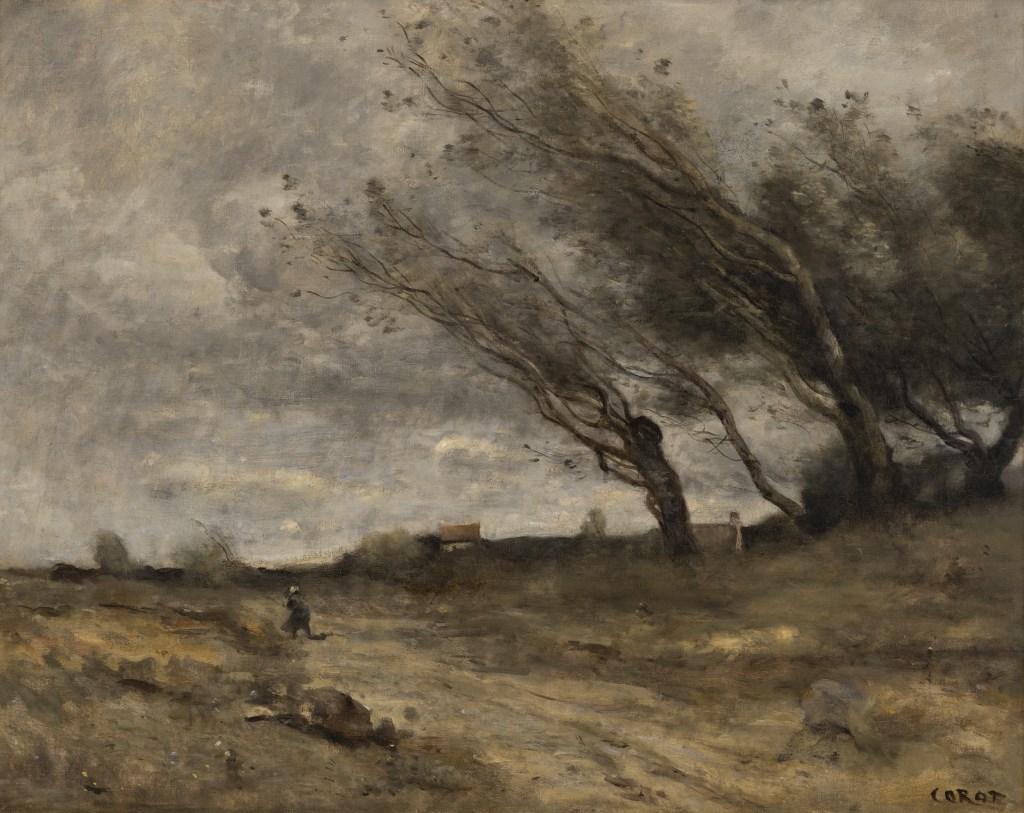}
  \quad
  \includegraphics[width=0.28\textwidth]{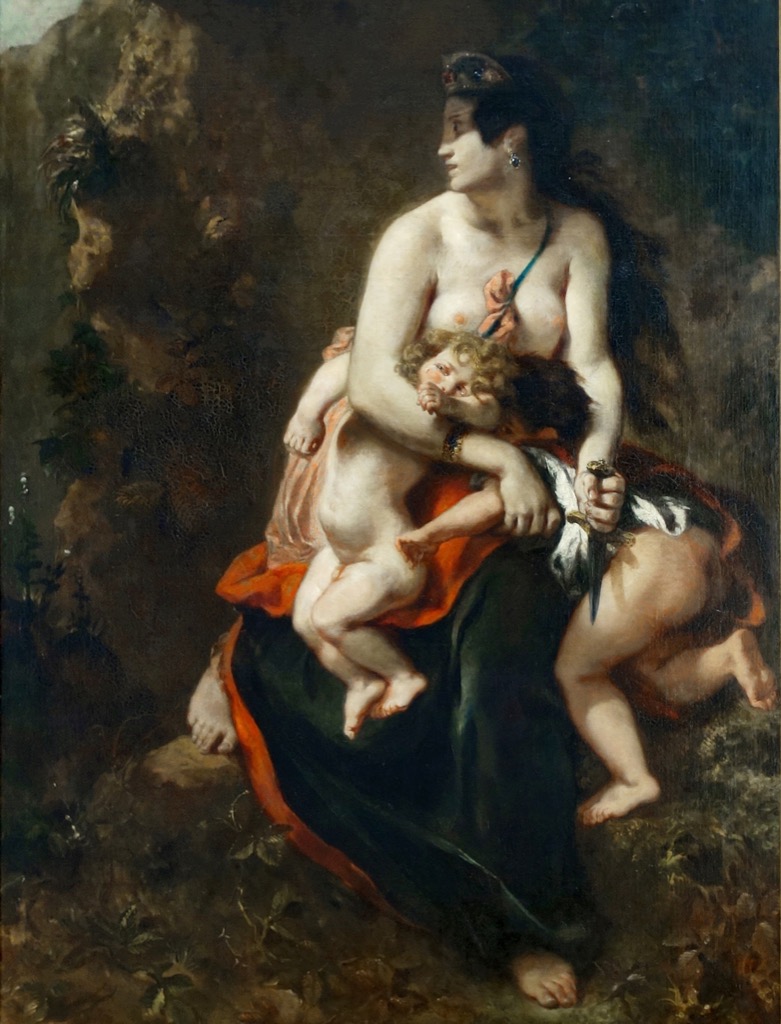}
  \quad
  \includegraphics[width=0.28\textwidth]{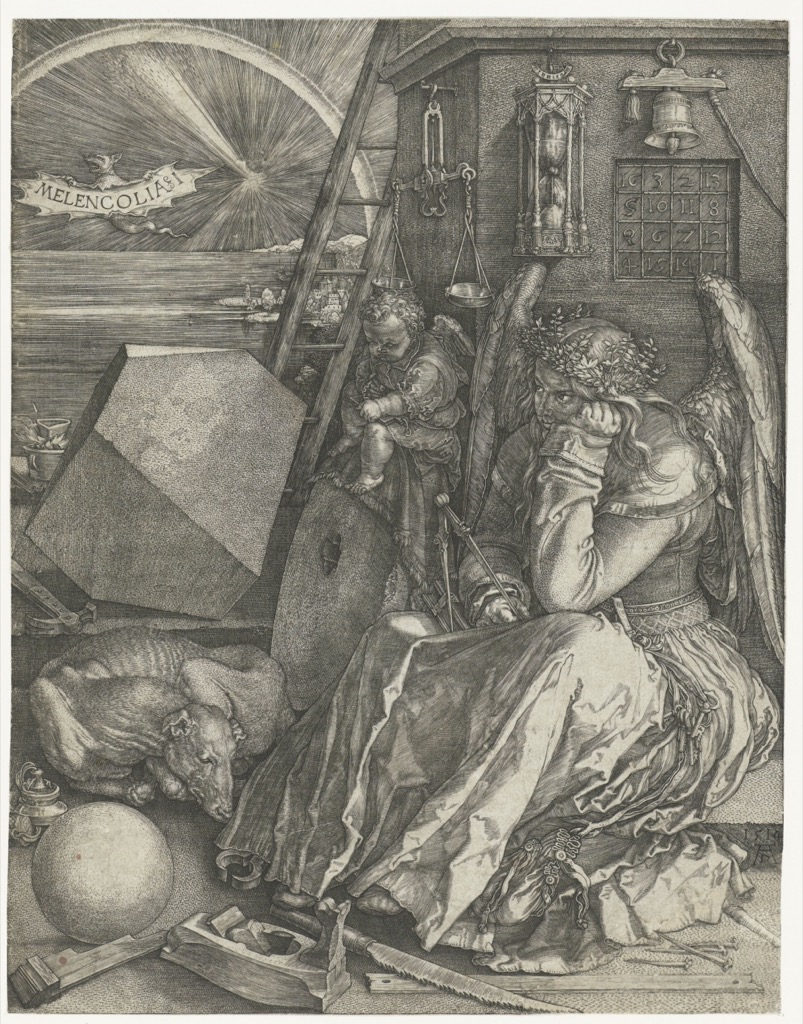}
  \caption{Examples of three difficult artworks. Left: The blast (Corot, Picture 15 in Appendix A). Center: Medea furious (Delacroix, Picture 21). Melancholia I (Dürer, Picture 27).}
    \label{fig:images}
\end{figure*}

\newcommand{\gr}[1]{\textcolor{DarkGreen}{#1}}
\newcommand{\pu}[1]{\textcolor{DarkMagenta}{#1}}
\newcommand{\red}[1]{\textcolor{Crimson}{#1}}

\begin{table*}
  \centering
  \tiny
  \setlength{\tabcolsep}{4pt}
\begin{tabular}{p{0.09\textwidth}p{0.20\textwidth}p{0.20\textwidth}p{0.22\textwidth}p{0.20\textwidth}}
    \toprule
    Prompt & Corot (\qwen) & Delacroix (\qwenawq) & Dürer (\qwen) & Dürer (\llava)  \\ \midrule
    2. What does this artwork show? &
This artwork depicts \gr{a landscape scene with a strong wind bending the trees}, creating a dynamic and dramatic effect. \gr{A solitary figure is seen walking along a path} in the foreground, adding a sense of scale and human presence to the vast, open environment. The \gr{sky is filled with clouds}, suggesting an overcast or stormy day. The painting captures the raw power of nature and the resilience of the human figure within it. 
                              &
                                This artwork depicts a classical scene featuring \gr{a nude woman} seated on a rocky surface, \pu{holding a child in her arms}. \gr{Another child is \red{playfully} interacting with her}, and the background includes lush foliage and a dark, shadowy setting, suggesting a mythological or allegorical theme. The composition and style are reminiscent of neoclassical or romantic art.
                                                   & This artwork is a detailed engraving that appears to depict a scene with \gr{symbolic elements}. It includes \gr{a figure in a state of repose}, possibly a reclining person or \red{a figure of death}, surrounded by various objects such as a globe, a saw, and other tools. There are also \red{two figures, one holding a baby, which could symbolize life or innocence, and another figure with wings, possibly an angel or a representation of time or fate}. The presence of \gr{the clock and the hourglass} suggests themes of time and mortality. [\dots] 
                                                     & The artwork depicts a scene with \gr{a woman sitting on the ground}, surrounded by various objects and figures that appear to be in \red{a state of disarray or chaos}. There is \gr{a ladder, a bell, a clock, and other items} scattered around, and the woman seems to be in a \pu{contemplative or distressed state}. The style is reminiscent of the Baroque period, known for its dramatic and intense compositions.
    \\ \midrule
    3. Does this artwork involve an emotion? &	Yes.  The artwork evokes a sense of \pu{tranquility and introspection}. The solitary figure walking through the landscape, combined with the gentle breeze and the expansive sky, suggests a \pu{peaceful moment of reflection}. The muted colors and the soft brushstrokes contribute to a \pu{calm and contemplative atmosphere}, which can be interpreted as an emotional response from the viewer.
& Yes.  This artwork involves emotion, as it depicts a \red{tender and intimate moment} between a woman and a child. The woman's posture and the way she holds the child suggest \red{a sense of care, love, and protection}. The overall composition, with its soft lighting and the natural setting, enhances the emotional depth of the scene, evoking \red{feelings of maternal love and serenity.}
                                                   &Yes.  The artwork depicts a scene with \pu{symbolic elements that can evoke emotions}. The presence of \red{the angel holding a baby}, the reclining figure, and the various objects around them suggest themes of \pu{mortality, divinity, and perhaps a sense of loss or contemplation}. The overall composition and the use of light and shadow can also contribute to an emotional atmosphere, inviting viewers to reflect on the deeper meanings behind the imagery.
                                                     & Yes, the artwork appears to involve an emotion. \gr{The central figure, which seems to be a woman, is depicted with a contemplative or possibly sorrowful expression}. \gr{Her posture and the way she holds her head suggests a moment of introspection or sadness}. The overall atmosphere of the scene, with the \pu{cluttered and somewhat chaotic environment, may also contribute to a sense of emotional distress or unease}.
    \\ \midrule
  4. Is the emotion shown by this artwork a positive or a negative one? &	\gr{Negative}  The painting depicts a stormy landscape with bent trees, suggesting strong winds and possibly rain, which can evoke feelings of \gr{unease, fear, or sadness} in viewers. The solitary figure in the distance might also contribute to a sense of \gr{isolation or vulnerability} [\dots]
& \red{**Positive**}  The artwork depicts a \red{serene and tender moment} between a woman and a child, with the woman holding the child close and \red{the child appearing content and secure}. The overall mood is one of \red{love, care, and tranquility}, which are positive emotions. [\dots]
& \gr{Negative}  The artwork depicts a \red{scene of death, with a figure lying on the ground}, possibly deceased, and a dog lying beside them. The presence of \red{an angel holding a baby} suggests a theme of loss or mourning, which typically conveys a negative emotional tone. &	\gr{Negative}
    \\ \midrule
    6. How is the emotion shown by this artwork expressed artistically? &	The emotion in this artwork is expressed through the \gr{dynamic and dramatic posture of the trees}, which appear to be bent and twisted by the wind, suggesting a sense of struggle and resilience. The \gr{muted color palette} and the \gr{overcast sky} contribute to a somber and contemplative mood, evoking feelings of solitude and introspection. The lone figure in the distance adds to the overall feeling of isolation and perhaps a yearning for connection or escape from the harshness of nature. The artist's use of \gr{light and shadow} further enhances the emotional depth [\dots]
                              & The artwork expresses emotion through a \gr{combination of compositional elements, color, and body language}. The central figure, a woman, is depicted with a \pu{serene and contemplative expression, her gaze directed outward, suggesting introspection or a sense of calm.} The use of soft, warm lighting highlights her form and the delicate features of the child she holds, \red{emphasizing tenderness and maternal love}. The dark, shadowy background contrasts with the illuminated figures, drawing attention to their interaction and \pu{creating a sense of intimacy}. [\dots]
                                                     & The emotion in this artwork is expressed through intricate details and \gr{symbolic elements}. The angelic figure\red{s, one holding a baby, convey a sense of tenderness and protection}. The reclining figure draped in a cloak suggests a state of rest or perhaps contemplation, while the presence of tools and \red{a globe hints at themes of craftsmanship and exploration}. The \gr{overall composition, with its detailed textures and symbolic objects} like the hourglass and bell, evokes a contemplative and somewhat somber mood, reflecting on themes of \gr{time, mortality, and the passage of life}.
                                                     & The emotion in this artwork is expressed through the use of a \pu{somber color palette}, the \gr{detailed rendering of the figures, and the overall composition}. The woman, dressed in a flowing gown, sits with \gr{her head in her hands, suggesting a state of sorrow or contemplation}. The surrounding environment, with its \pu{cluttered and chaotic elements, adds to the sense of distress}. The use of light and shadow, as well as the positioning of the figures, contributes to the overall mood of the piece. The artist has skillfully used these elements to convey a deep emotional impact.
    \\ \midrule
7. Does this artwork use a symbol to visualize emotion, and if it does, how?  &	Yes, the artwork uses the \pu{symbol of bent trees to visualize emotion}. The trees, leaning dramatically towards the left side of the canvas, suggest a strong wind or storm, evoking a sense of struggle and resilience. The figure in the foreground, seemingly small and distant, adds to the feeling of isolation and perhaps a contemplative mood amidst the powerful natural forces depicted.
& \red{Yes}, this artwork uses symbols to visualize emotion. The central figure, a woman, is depicted in a contemplative or melancholic pose, holding a child who appears to be in distress or fear. \pu{The child's expression and body language, such as clutching the woman's arm, symbolize} vulnerability and dependence. \pu{The dark, shadowy background and the woman's somber expression} further enhance the emotional tone, suggesting themes of sorrow, protection, or maternal concern. [\dots]
& \gr{Yes}, the artwork uses symbols to visualize emotion. \pu{The angel holding the baby conveys a sense of innocence and purity}, while the angel with the \red{laurel wreath and sword} suggests a \red{theme of victory or triumph over adversity}, possibly symbolizing the triumph of good over evil. The presence of the \gr{dog lying down might represent loyalty or rest}, adding another layer of emotional depth to the scene. The overall composition, with its detailed elements and symbolic figures, evokes a narrative that is rich in emotional and philosophical undertones.                       
& \gr{Yes}, the artwork uses a symbol to visualize emotion. The woman in the image is depicted with \red{a halo}, which is a common symbol in art to represent \red{holiness, purity, or sainthood}. This suggests that the woman is being portrayed as a virtuous or sacred figure, which can evoke feelings of reverence or admiration in the viewer. The halo's presence in the artwork serves to convey a sense of \pu{spiritual significance or moral excellence} associated with the character.  \\ \bottomrule
  \end{tabular}
  \caption{VLM generated descriptions for the difficult artworks. Green: correct, red: incorrect, purple: questionable.}
  \label{tab:descriptions}
\end{table*}

We discuss our findings structured by observation.
Figure~\ref{fig:images} shows three artworks to illustrate the
difficulties of the task\footnote{See Appendix for image source and
  licensing information.} and Table~\ref{tab:descriptions} the
corresponding VLM answers for the five most instructive questions. We
color-code relevant passages in the VLM output as correct (green),
incorrect (red), and questionable (purple).

\paragraph{A Hierarchy of Skills in Image Analysis.} All three models
demonstrate a good, sometimes even very good, ability to understand
the images. However, there are significant differences in answer
quality among the different question types, which roughly correlate
with the order of our questions: The initial questions are answered
better than those at the end. Q~1 (type of artwork) is almost always
answered correctly (\llava categorizes Hoepffner's black-and-white
photograph as a drawing). The content descriptions (Q~2) are also
mostly accurate: For Corot's 'Blast' (Fig.~\ref{fig:images} left
and Tab.~\ref{tab:descriptions}), the model correctly identifies the
motifs of the force of nature and the solitary person as well as the
effects they create.

Q~3 to~5, which concern the emotional content, are still
mostly answered correctly, but show a greater tendency toward evasive
or inconsistent answers. As an example, consider the Corot painting,
where Qwen initially speaks of a 'calm, contemplative atmosphere' and
a 'peaceful moment of reflection', but answers 'negative' to the
specific question about the polarity of the emotion and, with
reference to the forces of nature, also brings unease, fear and
vulnerability into play.

Q~6 concerning the artistic expression of emotions is answered
largely well. The models identify, mostly reliably, various levels of
expression of emotions, both in terms of content (composition,
depiction of specific objects, facial expressions and postures of
figures) and stylistic devices (color palette, brushwork). Overall,
facial expressions and postures are more precisely identified than
abstract forms which become the more accurate the more elements point
in the same direction. For example, in Camille Corot's 'Blast'
(Figure~\ref{fig:images} left), \qwen mentions the wind-blown trees
and the threatening sky against which the small human figure is seen,
in combination with the gloomy colors and the play of light and
shadow. Dynamic compositions tend to be recognized better than static
ones.

The answers to Q~7 concerning symbols are the least convincing. Here,
all models frequently mention aspects of the image that do not
represent symbols. Answers to questions 6 and 7 are often identical,
as in the landscape paintings by Caspar David Friedrich (Pictures
9/10, cf. Appendix~A), where, when asked about means of expression and
symbols, \qwen mentions soft and nuanced colors that evoke calm,
peace, and a contemplative mood. In some cases, symbolic objects that
do not appear in the paintings are also hallucinated. For example,
\llava calls the personification of Melancholy in Albrecht Dürer's
engraving an 'angel' (Figure~\ref{fig:images} right) because of her
wings and hallucinates a halo that she does not possess. This might be
due to the stylistic proximity to religious images, which often
feature halos.

Comparing the answers within the minimal pairs of similar artworks,
we also see a reasonable pattern: The answers are similar, making it
clear that the pictures are closely related, but sufficiently
dissimilar to pick out each picture's individuality.

In sum, we observe a continuum: Properties of images that can
characterized concretely at the visual level are captured more
reliably by the VLMs than properties requiring more
interpretation. However, even these are often inferred correctly at a
basic level. In our view, this ability indicates that the models were
exposed to (at least some) interpretive texts on art or art history
during training.

\paragraph{Overextension of the concepts emotion and symbol.}
Like LLMs, the VLMs consistently exhibit confirmation (or position)
bias, i.e., the tendency to answer 'yes' to yes/no questions
\cite{Echterhoff2024CognitiveBias}. This applies in particular to Qs~3
and 7 (presence of emotions and emotion symbols): The models answer
almost always affirmatively, even if this is wrong. Indeed, \llava
tends to avoid, in cases of doubt, a yes/no answer and simply lists
potential symbols. The only model that gives explicit negative answers
is \qwenawq ('This artwork does not use a symbol to visualize
emotion').

As the models continue such wrongly affirmative answers, they are
forced to interpret the terms 'emotion' and 'symbol' very
broadly. Regarding emotions, the models refer to a small portion of
the artworks in terms of emotions in the narrower sense of basic
emotions \cite{Ekman1999BasicEmotions} such as \textit{grief, fear}, or \textit{joy}. Many others are described by general affective states -- or
maybe aesthetic emotions \cite{Israeli1928Affective} -- such as
\textit{melancholy, loneliness}, or \textit{awe}. For others, the
models describe the emotional content with very abstract terms such as
\textit{contemplation, tranquility} and \textit{introspection}, with
only a tenuous connection to concrete visual
properties. \textit{Contemplation} in particular is used so frequently
-- by all models for more than half of all images -- that an
interpretation by reference to a (e.g., monastic) \textit{vita
  contemplativa} (a philosophical life style) is unconvincing. Given
that VLMs are presumably trained mostly on photographs, it appears
plausible that \textit{contemplation} characterizes the specific
aesthetic quality of the works of art, in contrast to the pictures of
contemporary reality that the models are more familiar with.

We see a similar situation with symbols: when clear symbols are
present in the images, they are often recognized, for example, the
book as a symbol of wisdom, the threatening sky and the thunderstorm
as symbols of conflict and threat, or the skull as a symbol of
mortality. \llava does a better job in this regard; see the discussion
below.

Particularly in less symbolic images, the models also tend to describe
any emotion-relevant properties of the images as symbols, such as
certain facial expressions or postures, the color palette, or the
contrast of light and shadow. In our opinion, these are not genuine
(emotion) symbols, since the emotional content here is part of the
primary meaning and not added through a secondary conventionalized
relationship. The models use the verb 'symbolize' very loosely to
simply mean 'express'. In other places, the models identify symbols
that are not conventionalized and therefore somewhat dubious. These
include the interpretation of a figure standing in water as a symbol
of loneliness (Whistler, Picture 33/34), or the interpretation of a
cypress tree as a symbol of timelessness (Böcklin, Picture
25/26). Finally, some clear symbols, such as the dagger in Medea's
hand (Figure~\ref{fig:images} center), are overlooked when they do not
fit a simple interpretation (see below).

\paragraph{Lack of Consistency.} The VLMs we consider here do not
possess a thought process independent of output generation. In
consequence, they are unable to reflect on the consistency of their
output \cite{marjanovic2025deepseekr1thoughtologyletsthink}. We
observe inconsistency both within answers and across answers. For
example, in justifying their answer to Q~3 (Is the artwork
emotional?), the models often already identify an emotion and the form
of its expression, which overlaps in content with Q~5 (specific
emotion) and Q~6 (emotion expression). In addition, there are semantic
relations between questions: In an ideal model, the answers to Q~4
(emotion polarity) and Q~5 (specific emotion) would always match.

We observe a correlation between consistency and the emotion's
intensity (Q~8): the clearer and more intense the emotion depicted,
the more consistently the models answer, similar to human behavior
\cite{troiano21:_entan_annot_confid_emotion_inten}. For images with
subjective, weak, or ambivalent emotions, we often see contradictory
answers to the various questions. Camille Corot's 'Blast'
(Figure~\ref{fig:images}, left) falls into this category, as mentioned
above.  The models also frequently use hedging language ('or',
'perhaps even'). In principle, this might be an informative strategy,
if it allowed the user to recognize model uncertainty. Currently,
however, such hedge expressions are not used with sufficient
reliability by the models: not all uncertain statements are marked as
such, and even attributions that we consider unambiguous are
repeatedly weakened. Thus, inconsistency is evident at this level as
well.

\paragraph{Easier and Harder Artwork Categories.} When comparing
artworks from different periods, it is striking that the models
produce the best results for representational artworks (showing
recognizable objects), independently of the artistic
technique. Examples include the \textit{pietà} sculptures as well as
paintings by Corot, Seurat, and Van Gogh. In comparison, the models
struggle with three other categories of artworks: abstract images,
allegories, and highly contextualized artworks.

There are three highly abstract images in our sample: two Deluge
paintings by William Turner, two water paintings by James McNeill
Whistler (Pictures 33/34), and two sections of a Mark Rothko mural
(Pictures 35/36). These paintings primarily convey moods through
colors and shapes, containing few concrete object. The models do
recognize these moods, but due to semantic underdetermination, they
tend to detect both positive and negative emotions, thus, as discussed
above, offering an inconsistent interpretation. Formulations such as
'depending on the context' or 'depending on the viewer' are also
offered.

The category of allegories is exemplified in our sample by the two
allegories on melancholy by Albrecht Dürer (Figure \ref{fig:images}
right) and Giovanni Benedetto Castiglione (Pictures 27/28). The models
are clearly unfamiliar with the genre and cannot cope with the fact
that (almost) all of the objects in the painting are to be understood
symbolically. As shown in Table~\ref{tab:descriptions} (right), the
models mention 'chaos' or describe the painting as 'overloaded'.
Indeed, recognizing the image content also seems more difficult than
in other categories, and \qwen, in particular, hallucinates several
related objects (including an angel holding a baby and a clock). As a
result, the models' explanations of the emotions remain vague and
contradictory, even though the general negative polarity is correctly
recognized.

The third difficult category consists of artworks that require
background knowledge for interpretation, such as Eugène Delacroix's
painting of Medea (Figure~\ref{fig:images} center) from Greek
mythology who murders her children out of spite. \qwen recognizes the
relevant content (mother, children, sword, cave), the stylistic
devices (contrasting colors), and also the tension depicted. In the
overall interpretation, however, the model focuses on the main objects
of mother and children and concludes that the image symbolizes
'balance and tenderness\dots the emotional depth of motherly
love'. This may also be related to the fact that the model was
(presumably) forbidden from verbalizing violent content due to its
safety alignment; however, the problem also arises in other cases.

The difficult cases have in common that the emotional
interpretation cannot be derived from the co-occurrence of a few
objects associated with a conventionalized meaning: Either because
there are no objects (abstract images), because there are too many
(symbolic images), or because the conventional interpretation is
overridden by a specific context (mythological images). This
observation suggests that good recognition of objects and their
configurations is still important for sound image interpretation. This
seems plausible: the images the VLMs saw during training are likely
only a fraction of works of art, and a much larger proportion are
photographs, often of a journalistic nature. In these photographs, the
meaning is primarily derived from the objects and only to a small
extent from other properties. Furthermore, such popular or
journalistic images usually operate with basic anthropological
constellations (love, grief, relationships, violence) and less with
complex symbols, as they are designed for broad understanding. From
this perspective, it is even surprising that the VLMs are relatively
successful in recognizing the stylistic and formal properties of the
artworks. Again, this indicates the presence of art (historical)
materials in the training data.

\paragraph{Differences among VLMs.} The three models are comparable in
terms of the general quality of their analysis, despite their
differences in training and model sizes. They have different strengths
and weaknesses, though. Table~\ref{tab:descriptions} (right) shows the
output of \llava and \qwenawq for the same picture for direct
comparison.

The Qwen models appear to have seen more texts in their training that
deal with visual arts specifically. As a consequence, their image
descriptions read more professionally and often convey the content
better. For example, in the landscape paining 'Morning on the
Riesengebirge' (Caspar David Friedrich, Picture 10), the gentle
gradation from warm yellow to cooler blue and the staggered mountain
ranges are described in terms of a feeling of depth and expanse that
invites the viewer to reflect on the grandeur of nature, while the
sparse vegetation and the deserted landscape evoke feelings of
loneliness and introspection: there is (almost) nothing to add to this
from an art historical perspective.  The model also recognizes other
artistic means that do not operate as symbols, such as the half-open
doors in Vilhelm Hammershøi's work (Pictures 31/32) that create
tension. The Qwen models also has significantly more extensive factual
knowledge: \qwen directly recognizes some artists (Van Gogh), styles
(Pointillism), and categories of artworks (\textit{pietà}, Pictures
3--6). It derives emotional values directly from precisely recognized
religious iconography (\textit{pietà} = grief, empathy, maternal
love). This knowledge also leads to correct intercultural
interpretation of expressive values (Indian lady, 18th century,
Picture 30).

The danger is that models rely too heavily on their knowledge of
specific images. In fact, the larger \qwenawq model mistakes a numbers
of artworks for different but related ones, for example claiming that
George Seurat's 'Circus sideshow' (Picture~7) was his (better-known)
'A Sunday Afternoon on the Island of La Grande Jatte'. Consequently,
the description of the painting is a mixture of what is seen in the
actual image and what the model knows about the other one ('The
figures in the painting are engaged in various activities\dots'). In
that sense, the larger model's capacity for better memorization is in
fact a liability. On the upside, however, the larger \qwenawq -- as
discussed above -- is also the only one to break the confirmation bias
and state correctly that some pictures are not symbolic.

In comparison, \llava has less domain knowledge: the model only
roughly recognizes styles and no artists. The model tends to have more
difficulty interpreting complex images than the Qwen models, resorting
more often to vague or attenuated expressions. A surprising strength
of \llava is its ability to recognize symbols in images and assign
emotions better than Qwen. For example, \llava recognizes the veil in
Marta Hoepffner's 'Mourning' photograph (Picture 24); the book as a
symbol of knowledge and wisdom in several images; the abyss as a
symbol of threat; individual small figures as symbols of loneliness;
and the cross and crown of thorns as symbols of the Passion -- the
latter even though the model has no concept of \textit{pietà}. The
model thus demonstrates a knowledge of affective symbols from
different eras and cultures.

\section{Discussion}

Our case study investigated how well current vision-language models
(VLMs) can be used to interpret the emotional content of artworks,
presenting a dataset of 38 images to three current VLMs. By using a
qualitative evaluation approach based on expert judgments, we are able
to obtain a detailed profile of the VLMs' capabilities. Our results
show that VLMs can recognize the content of artworks well, often also
the emotions they depict and how they are expressed -- largely
independent of the type of artwork, their historical period, and their
style. This indicates that VLMs implicitly incorporate a substantial
amount of art historical knowledge on which they can build. The Qwen
models demonstrate significantly more detailed knowledge and a better
command of the domain language than \llava.

However, VLMs continue to exhibit the typical limitations of LMs: they
base their interpretations of deeper levels of meaning on observations
of simple surface patterns (such as the presence of certain objects or
stylistic devices) and their conventionalized meaning. This explains
their ability to generalize to artworks, but fails precisely when the
artist creates something novel by transcending conventionalized
patterns or at least using them ambivalently. Similarly, when the
patterns become too complex, as in symbolist paintings, the models
fail, since globally coherent interpretations would require complex
inferences \cite{Shen2024CulturalCommonsense}. Given these
observations, it is not surprising that the VLMs -– especially the
Qwen models -– struggle with the reliable recognition of emotion
symbols: The interpretation of such symbols typically arises from a
long chain of inferences that determine the meaning of a symbol by
intersecting the set of possible symbols, the set of interpretations
of these symbols, and the set of meanings of the entire artwork.

The strength of VLMs our study identifies in dealing with
conventionalized aspects of artwork already lends itself to possible
concrete applications. For example, VLMs could be used in extending
large catalogs of artworks with short descriptive text for better
indexing. Similarly, models also seem to be sufficiently powerful for
the large field of audio descriptions in e-publishing, converting
images into text, for example to improve accessibility. In either
case, there is still a role for experts to review and correct these
results.

\section{Limitations and Future Work}

Our analysis is a case study and only considered a limited number of
artworks (38) and of VLMs (3). The artworks were all drawn from the
classical art history canon, including only one non-Western artwork
(Picture 30) and only two artworks by women (Picture 23 and 24). 

Further, our analysis of the VLM outputs follows methodological
practice in art history rather than AI. It adopts a scalable reading
approach, largely qualitative in nature, that does not scale well: it
could not be extended easily to larger samples of artworks (to achieve
a better representativeness), additional questions (to apply our
approach to artwork properties other than emotions) or a larger number
of raters (to minimize the impact of personal bias).

For these reasons, one important avenue for future research is the
identification of evaluation approaches that combine the advantages of
Ozaki et al.'s (2024) fully automatic quantitative analysis with our
qualitative, more detailed analysis. A second such avenue is the
fine-tuning of VLMs for the analysis of emotions in artwork. This can
be hoped to improve the general ability of VLMs to recognize emotional
content. Also, it would ideally curb the models' verbosity, which
complicates their use for tasks like audio description generation (as
mentioned above) and the field of cultural heritage in general. As
usual, the challenge remains to find suitable supervision.

\bibliography{main.bib}

@inproceedings{troiano21:_entan_annot_confid_emotion_inten,
  author = {Troiano, Enrica and Padó, Sebastian and Klinger, Roman},
  booktitle = {Proceedings of the EACL WASSA workshop},
  pages = {50--61},
  title = {Emotion Ratings: How Intensity, Annotation Conﬁdence and Agreements are Entangled},
  url = {https://www.aclweb.org/anthology/2021.wassa-1.5},
  year = 2021
}

@inproceedings{hayashi-etal-2024-towards,
    title = "Towards Artwork Explanation in Large-scale Vision Language Models",
    author = "Hayashi, Kazuki  and
      Sakai, Yusuke  and
      Kamigaito, Hidetaka  and
      Hayashi, Katsuhiko  and
      Watanabe, Taro",
    editor = "Ku, Lun-Wei  and
      Martins, Andre  and
      Srikumar, Vivek",
    booktitle = "Proceedings of the 62nd Annual Meeting of the Association for Computational Linguistics (Volume 2: Short Papers)",
    month = aug,
    year = "2024",
    address = "Bangkok, Thailand",
    publisher = "Association for Computational Linguistics",
    url = "https://aclanthology.org/2024.acl-short.65/",
    doi = "10.18653/v1/2024.acl-short.65",
    pages = "705--729"
}

@inproceedings{ozaki-etal-2025-towards,
    title = "Towards Cross-Lingual Explanation of Artwork in Large-scale Vision Language Models",
    author = "Ozaki, Shintaro  and
      Hayashi, Kazuki  and
      Sakai, Yusuke  and
      Kamigaito, Hidetaka  and
      Hayashi, Katsuhiko  and
      Watanabe, Taro",
    editor = "Chiruzzo, Luis  and
      Ritter, Alan  and
      Wang, Lu",
    booktitle = "Findings of the Association for Computational Linguistics: NAACL 2025",
    month = apr,
    year = "2025",
    address = "Albuquerque, New Mexico",
    publisher = "Association for Computational Linguistics",
    url = "https://aclanthology.org/2025.findings-naacl.209/",
    doi = "10.18653/v1/2025.findings-naacl.209",
    pages = "3773--3809",
    ISBN = "979-8-89176-195-7"
}

@incollection{saPereira2019mixed,
  author    = {de Sá Pereira, Moacir P.},
  title     = {Mixed Methodological Digital Humanities},
  booktitle = {Debates in the Digital Humanities 2019},
  editor    = {Gold, Matthew K. and Klein, Lauren F.},
  publisher = {University of Minnesota Press},
  address   = {Minneapolis},
  year      = {2019},
}

@misc{marjanovic2025deepseekr1thoughtologyletsthink,
      title={{DeepSeek-R1} Thoughtology: Let's think about {LLM} Reasoning}, 
      author={Sara Vera Marjanović and Arkil Patel and Vaibhav Adlakha and Milad Aghajohari and Parishad BehnamGhader and Mehar Bhatia and Aditi Khandelwal and Austin Kraft and Benno Krojer and Xing Han Lù and Nicholas Meade and Dongchan Shin and Amirhossein Kazemnejad and Gaurav Kamath and Marius Mosbach and Karolina Stańczak and Siva Reddy},
      year={2025},
      eprint={2504.07128},
      archivePrefix={arXiv},
      primaryClass={cs.CL},
      url={https://arxiv.org/abs/2504.07128}, 
}

@Book{cassirer,
  author = 	 {Ernst Cassirer},
  title = 	 {Philosophie der symbolischen Formen},
  publisher = 	 {Kindler},
  volume = {1: Die Sprache},
  year = 	 {1923},
  address = 	 {Weimar}}

@article{Naeslund01102020,
author = {Anna Näslund and Amanda Wasielewski},
title = {Cultures of Digitization: A Historiographic Perspective on Digital Art History},
journal = {Visual Resources},
volume = {36},
number = {4},
pages = {339--359},
year = {2020},
publisher = {Routledge},
doi = {10.1080/01973762.2021.1928864},
}

@article{Israeli1928Affective,
  author    = {Israeli, Nathan},
  title     = {Affective reactions to painting reproductions: A study in the psychology of esthetics},
  journal   = {Journal of Applied Psychology},
  volume    = {12},
  number    = {1},
  pages     = {125--139},
  year      = {1928},
  url       = {https://doi.org/10.1037/h0070445}
}

@inproceedings{Antol2015VQA,
  author    = {Antol, Stanislaw and Agrawal, Aishwarya and Lu, Jiasen and Mitchell, Margaret and Batra, Dhruv and Zitnick, C. Lawrence},
  title     = {{VQA}: Visual Question Answering},
  booktitle = {Proceedings of the IEEE International Conference on Computer Vision},
  year      = {2015},
  pages     = {2425--2433},
  address   = {Santiago, Chile}
}

@book{Barthes1977-BARIMT-3,
  author    = {Barthes, Roland},
  title     = {Image, Music, Text: Essays},
  year      = {1977},
  publisher = {Hill and Wang / Fontana Press}}

@article{Bai2025Qwen25VL,
  author    = {Bai, Shuai and Chen, Keqin and Liu, Xuejing and Wang, Jialin and Ge, Wenbin and Song, Sibo and Dang, Kai and Wang, Peng and Wang, Shijie and Tang, Jun and Zhong, Humen and Zhu, Yuanzhi and Yang, Mingkun and Li, Zhaohai and Wan, Jianqiang and Wang, Pengfei and Ding, Wei and Fu, Zheren and Xu, Yiheng and Ye, Jiabo and Zhang, Xi and Xie, Tianbao and Cheng, Zesen and Zhang, Hang and Yang, Zhibo and Xu, Haiyang and Lin, Junyang},
  title     = {Qwen2.5-{VL} Technical Report},
  journal   = {arXiv preprint arXiv:2502.13923}, 
  year      = {2025}
}

@inproceedings{Brown2020GPT3,
  author    = {Brown, Tom B. and Mann, Benjamin and Ryder, Nick and Subbiah, Melanie and Kaplan, Jared and Dhariwal, Prafulla and Neelakantan, Arvind and Shyam, Pranav and Sastry, Girish and Askell, Amanda and Agarwal, Sandhini and Herbert-Voss, Ariel and Krueger, Gretchen and Henighan, Tom and Child, Rewon and Ramesh, Aditya and Ziegler, Daniel M. and Wu, Jeffrey and Winter, Clemens and Hesse, Christopher and Chen, Mark and Sigler, Eric and Litwin, Mateusz and Gray, Scott and Chess, Benjamin and Clark, Jack and Berner, Christopher and McCandlish, Sam and Radford, Alec and Sutskever, Ilya and Amodei, Dario},
  title     = {Language Models are Few-Shot Learners},
  booktitle = {Advances in Neural Information Processing Systems},
  year      = {2020}
}

@incollection{Dieckmann2010Prometheus,
  author    = {Dieckmann, Lisa},
  title     = {Prometheus: The Distributed Digital Image Archive for Research and Education},
  booktitle = {L'Art et la Mesure},
  editor    = {Joyeux-Prunel, Béatrice},
  publisher = {Éditions Rue d'Ulm},
  year      = {2010},
  doi       = {10.4000/books.editionsulm.8647},
  url       = {https://doi.org/10.4000/books.editionsulm.8647}
}

@article{Kim2019SurveyLit,
  author    = {Kim, Evgeny and Klinger, Roman},
  title     = {A Survey on Sentiment and Emotion Analysis for Computational Literary Studies},
  journal   = {Zeitschrift für digitale Geisteswissenschaften},
  volume    = {4},
  pages     = {1--23},
  address   = {Wolfenbüttel},
  year      = {2019},
  doi       = {10.17175/2019_008_v2},
  url       = {https://doi.org/10.17175/2019_008_v2}
}

@inproceedings{Echterhoff2024CognitiveBias,
  author    = {Echterhoff, Jessica M. and Liu, Yao and Alessa, Abeer and McAuley, Julian and He, Zexue},
  title     = {Cognitive Bias in Decision-Making with {LLMs}},
  booktitle = {Findings of the Association for Computational Linguistics: EMNLP 2024},
  year      = {2024},
  pages     = {12640--12653},
  address   = {Miami, Florida, USA},
  publisher = {Association for Computational Linguistics}
}

@incollection{Ekman1999BasicEmotions,
  author    = {Ekman, Paul P.},
  title     = {Basic Emotions},
  booktitle = {The Handbook of Cognition and Emotion},
  editor    = {Dalgleish, T. and Power, T.},
  publisher = {John Wiley \& Sons},
  address   = {Sussex, U.K.},
  pages     = {45--60},
  year      = {1999}
}

@book{Frevert2011Gefuehlswissen,
  author    = {Frevert, Ute and Scheer, Monique and Schmidt, Anne and Eitler, Pascal and Hitzer, Bettina and Verheyen, Nina and Gammerl, Benno and Bailey, Christian and Pernau, Margrit},
  title     = {Gefühlswissen. Eine lexikalische Spurensuche in der Moderne},
  publisher = {Campus},
  address   = {Frankfurt am Main},
  year      = {2011}
}

@article{Huang2025HallucinationSurvey,
  author    = {Huang, Lei and Yu, Weijiang and Ma, Weitao and Zhong, Weihong and Feng, Zhangyin and Wang, Haotian and Chen, Qianglong and Peng, Weihua and Feng, Xiaocheng and Qin, Bing and Liu, Ting},
  title     = {A Survey on Hallucination in Large Language Models: Principles, Taxonomy, Challenges, and Open Questions},
  journal   = {ACM Transactions on Information Systems},
  volume    = {43},
  number    = {2},
  pages     = {42:1--42:55},
  year      = {2025}
}

@inproceedings{Liesenfeld2023OpeningChatGPT,
  author    = {Liesenfeld, Andreas and Lopez, Alianda and Dingemanse, Mark},
  title     = {Opening up {ChatGPT}: Tracking openness, transparency, and accountability in instruction-tuned text generators},
  booktitle = {Proceedings of the 5th International Conference on Conversational User Interfaces},
  publisher = {Association for Computing Machinery},
  address   = {New York, NY, USA},
  year      = {2023},
  pages     = {Article 47, 1--6}
}

@article{Liu2024HallucinationVLM,
  author    = {Liu, Hanchao and Xue, Wenyuan and Chen, Yifei and Chen, Dapeng and Zhao, Xiutian and Wang, Ke and Hou, Liping and Li, Rongjun and Peng, Wei},
  title     = {A Survey on Hallucination in Large Vision-Language Models},
  journal   = {arXiv preprint arXiv:2402.00253},
  year      = {2024}
}

@inproceedings{MLSYS2024_42a452cb,
 author = {Lin, Ji and Tang, Jiaming and Tang, Haotian and Yang, Shang and Chen, Wei-Ming and Wang, Wei-Chen and Xiao, Guangxuan and Dang, Xingyu and Gan, Chuang and Han, Song},
 booktitle = {Proceedings of Machine Learning and Systems},
 editor = {P. Gibbons and G. Pekhimenko and C. De Sa},
 pages = {87--100},
 title = {{AWQ}: Activation-aware Weight Quantization for On-Device {LLM} Compression and Acceleration},
 url = {https://proceedings.mlsys.org/paper_files/paper/2024/file/42a452cbafa9dd64e9ba4aa95cc1ef21-Paper-Conference.pdf},
 volume = {6},
 year = {2024}
}

@inproceedings{Liu2023VisualInstruction,
  author    = {Liu, Haotian and Li, Chunyuan and Wu, Qingyang and Lee, Yong Jae},
  title     = {Visual Instruction Tuning},
  booktitle = {Advances in Neural Information Processing Systems},
  year      = {2023}
}

@inproceedings{Lu2019ViLBERT,
  author    = {Lu, Jiasen and Batra, Dhruv and Parikh, Devi and Lee, Stefan},
  title     = {{ViLBERT}: Pretraining Task-Agnostic Visiolinguistic Representations for Vision-and-Language Tasks},
  booktitle = {Advances in Neural Information Processing Systems},
  year      = {2019}
}

@article{Matt2011EmotionHistory,
  author    = {Matt, Susan J.},
  title     = {Current Emotion Research in History: Or, Doing History from the Inside Out},
  journal   = {Emotion Review},
  volume    = {3},
  number    = {1},
  pages     = {117--124},
  year      = {2011}
}

@article{nandwani21,
	author = {Nandwani, Pansy and Verma, Rupali},
	date = {2021/08/28},
	doi = {10.1007/s13278-021-00776-6},
	id = {Nandwani2021},
	isbn = {1869-5469},
	journal = {Social Network Analysis and Mining},
	publisher = {SpringerNature},
	number = {1},
	pages = {81},
	title = {A review on sentiment analysis and emotion detection from text},
	url = {https://doi.org/10.1007/s13278-021-00776-6},
	volume = {11},
	year = {2021}}

@book{Plamper2012GeschichteGefuehl,
  author    = {Plamper, Jan},
  title     = {Geschichte und Gefühl. Grundlagen der Emotionsgeschichte},
  publisher = {Siedler},
  address   = {München},
  year      = {2012}
}

@article{Qin2025MultilingualLLM,
  author    = {Qin, Libo and Chen, Qiguang and Zhou, Yuhang and Chen, Zhi and Li, Yinghui and Liao, Lizi and Li, Min and Che, Wanxiang and Yu, Philip S.},
  title     = {A survey of multilingual large language models},
  journal   = {Patterns},
  volume    = {6},
  number    = {1},
  pages     = {101118},
  year      = {2025}
}

@inproceedings{Radford2021CLIP,
  author    = {Radford, Alec and Kim, Jong Wook and Hallacy, Chris and Ramesh, Aditya and Goh, Gabriel and Agarwal, Sandhini and Sastry, Girish and Askell, Amanda and Mishkin, Pamela and Clark, Jack and Krueger, Gretchen and Sutskever, Ilya},
  title     = {Learning Transferable Visual Models From Natural Language Supervision},
  booktitle = {Proceedings of the 38th International Conference on Machine Learning},
  year      = {2021}
}

@article{Rosenwein2010ProblemsEmotions,
  author    = {Rosenwein, Barbara H.},
  title     = {Problems and Methods in the History of Emotions},
  journal   = {Passions in Context},
  volume    = {1},
  volume = {1},
  pages = {1--32},
  year      = {2010}
}

@inproceedings{Shen2024CulturalCommonsense,
  author    = {Shen, Siqi and Logeswaran, Lajanugen and Lee, Moontae and Lee, Honglak and Poria, Soujanya and Mihalcea, Rada},
  title     = {Understanding the Capabilities and Limitations of Large Language Models for Cultural Commonsense},
  booktitle = {Proceedings of the 2024 Conference of the North American Chapter of the Association for Computational Linguistics: Human Language Technologies},
  volume    = {1},
  pages     = {5668--5680},
  address   = {Mexico City, Mexico},
  publisher = {Association for Computational Linguistics},
  year      = {2024}
}

@article{Stearns1985Emotionology,
  author    = {Stearns, Peter N. and Stearns, Carol Z.},
  title     = {Emotionology. Clarifying the History of Emotions and Emotional Standards},
  journal   = {The American Historical Review},
  volume    = {90},
  number    = {4},
  pages     = {813--830},
  year      = {1985}
}

@incollection{Tan2000EmotionArt,
  author    = {Tan, Ed S.},
  title     = {Emotion, Art, and the Humanities},
  booktitle = {Handbook of Emotions},
  editor    = {Lewis, M. and Haviland-Jones, J. M.},
  edition   = {2},
  publisher = {Guilford Press},
  address   = {New York},
  pages     = {116--134},
  year      = {2000}
}

@inproceedings{Vaswani2017Attention,
  author    = {Vaswani, Ashish and Shazeer, Noam and Parmar, Niki and Uszkoreit, Jakob and Jones, Llion and Gomez, Aidan N. and Kaiser, Lukasz and Polosukhin, Illia},
  title     = {Attention Is All You Need},
  booktitle = {Advances in Neural Information Processing Systems},
  year      = {2017}
}

@inproceedings{Wang2024CoT,
  author    = {Wang, Xuezhi and Zhou, Denny},
  title     = {Chain-of-Thought Reasoning Without Prompting},
  booktitle = {Advances in Neural Information Processing Systems},
  year      = {2024}
}

@article{Weitin2017ScalableReading,
  author    = {Weitin, Thomas},
  title     = {Scalable Reading},
  journal   = {Zeitschrift für Literaturwissenschaft und Linguistik},
  volume    = {47},
  pages     = {1--6},
  year      = {2017}
}

@article{Zhang2024VLSurvey,
  author    = {Zhang, Jingyi and Huang, Jiaxing and Jin, Sheng and Lu, Shijian},
  title     = {Vision-Language Models for Vision Tasks: A Survey},
  journal   = {IEEE Transactions on Pattern Analysis and Machine Intelligence},
  volume    = {46},
  number    = {8},
  pages     = {5625--5644},
  year      = {2024}
}

\appendix

\section{Image List}

The following list covers artists, image titles, preprocessing steps
(where pertinent) and licensing information (for the images shown in
Fig.~1).

\begin{supertabular}{rp{0.8\columnwidth}}
\toprule
\textbf{No.} & \textbf{Image}  \\
\midrule
  1 & Charles Le Brun: \textit{Fear (D)} \\
    & [Preprocessing: Writing masked] \\
  2 & Anne-Louis Girodet: \textit{Scene from a deluge (P)}  [Preprocessing: Writing masked] \\
  3 & Unknown artist: \textit{Pietà Röttgen (S)} \\
  4 & Unknown artist: \textit{Pietà 1875 (S)} \\
  5 & Giovanni de Fondulis: \textit{Madonna on throne with child (S)} \\
  6 & Michelangelo: \textit{Pietà (S)} \\
  7 & Georges Seurat: \textit{Circus sideshow (P)} \\
  8 & Georges Seurat: \textit{The circus (P)}  \\
  9 & Caspar David Friedrich: \textit{The Monk by the Sea (P)} \\
  10& Caspar David Friedrich: \textit{Morning on the Riesengebirge (P)} \\
  11& William Turner: \textit{Shade and Darkness – the evening of the deluge (P)} \\
  12&William Turner: \textit{Light and colour (Goethe's theory) -- the morning after the deluge -- Moses writing the book of genesis (P)} \\
  13 & Nicolas Poussin: \textit{Landscape during thunderstorm with Pyramus and Thisbe (P)} \\
  14 & Nicolas Poussin: \textit{Landscape with Saint Jerome (P)} \\
  15 & Camille Corot: \textit{The blast (P)} \\
  & Shown in Fig.~1 (left). License: CC BY \\
  & Source: \url{https://musees-reims.fr} \\
  16 & Camille Corot: \textit{Recollection of Mortefontaine (P)} \\
  17 & Claude Lorrain: \textit{Harbour scene at sunset (P)} \\
  18 & Claude Lorrain: \textit{Harbour scene with rising sun (P)}\\
  19 & Caspar David Friedrich: \textit{View of Arkona with rising moon (D)} \\
  20 & Caspar David Friedrich: \textit{View of Arkona by moonlight (D)} \\
  21 & Eugène Delacroix: \textit{Medea furious (P)} \\
  & Shown in Fig.~1 (center). License: PDM \\
  & Source: Wikipedia Commons \\
  22 & William Wetmore Story: \textit{Medea (S)} \\
  23 & Clara von Rappard: \textit{In Trauer (P)} \\
  24 & Marta Hoepffner: \textit{Ausdrucksstudie Trauer (F)} \\
  25 & Arnold Böcklin: \textit{Villa by the sea (P)} \\
  26 & Arnold Böcklin: \textit{Rest on the flight into Egypt (P)} [Preprocessing: Left part of picture (figures) masked] \\
  27 & Albrecht Dürer: \textit{Melancholia I (D)} \\
  & [Preprocessing: Writing masked] \\
  & Shown in Fig.~1 (right). License: CC0 \\
  & Source: Wikipedia Commons \\
  28 & Giovanni Benedetto Castiglione: \textit{Melancholia (D)}  \\
  29 & Henri Matisse: \textit{Odalisque (P)} \\
  30 & Unknown artist: \textit{Melancholy Courtesan (P)}  \\
  31 & Vilhelm Hammershøi: \textit{Interior. Study in sunlight (P)} \\
  32 & Vilhelm Hammershøi: \textit{White doors / Open doors (P)} \\
  33 & James McNeill Whistler: \textit{Nocturne: Blue and silver -- Cremone lights (P)} \\
  34 & James McNeill Whistler: \textit{Nocturne: Blue and silver -- Chelsea (P)} \\
  35 & Mark Rothko: \textit{Mural, Section 5 (P)} \\
  36 & Mark Rothko: \textit{Untitled [Black on maroon (Seagram Mural)] (P)} \\
  37 & Vincent van Gogh: \textit{Van Gogh's chair (P)} \\
  38 & Vincent van Gogh: \textit{Gauguin's chair (P)}  \\
  \bottomrule
\end{supertabular}
\vspace{\baselineskip}

\noindent Remarks:
\begin{itemize}
\item The three images from Figure~1 are numbers 15, 21, and
  27. Details see there.
\item Artwork types are abbreviated as follows: P = painting, S = sculpture, F = photography, D = drawing
\item Complete information on pictures is available at \url{https://doi.org/10.5281/zenodo.17668625}
\end{itemize}

\end{document}